\documentclass{article}
\usepackage{arxiv}
\usepackage{graphicx} 
\usepackage{multirow}
\usepackage{hyperref}
\usepackage{subcaption}
\usepackage{todonotes}

\title{LLaMA-NAS: Efficient Neural Architecture Search for Large Language Models}

\begin{document}

\author{
    Anthony Sarah \\
    \small{Intel Labs, Intel Corporation} \\
    \small{anthony.sarah@intel.com} \\\And
    Sharath Nittur Sridhar \\
    \small{Intel Labs, Intel Corporation} \\
    \small{sharath.nittur.sridhar@intel.com} \\\And
    Maciej Szankin \\
    \small{Intel Labs, Intel Corporation} \\
    \small{maciej.szankin@intel.com} \\\And
    Sairam Sundaresan \\
    \small{Intel Labs, Intel Corporation} \\
    \small{sairam.sundaresan@intel.com} \\\
}

\maketitle

\begin{abstract}

The abilities of modern large language models (LLMs) in solving natural language processing, complex reasoning, sentiment analysis and other tasks have been extraordinary which has prompted their extensive adoption. Unfortunately, these abilities come with very high memory and computational costs which precludes the use of LLMs on most hardware platforms. To mitigate this, we propose an effective method of finding Pareto-optimal network architectures based on LLaMA2-7B using one-shot NAS. In particular, we fine-tune LLaMA2-7B only once and then apply genetic algorithm-based search to find smaller, less computationally complex network architectures. We show that, for certain standard benchmark tasks, the pre-trained LLaMA2-7B network is unnecessarily large and complex. More specifically, we demonstrate a 1.5x reduction in model size and 1.3x speedup in throughput for certain tasks with negligible drop in accuracy. In addition to finding smaller, higher-performing network architectures, our method does so more effectively and efficiently than certain pruning or sparsification techniques. Finally, we demonstrate how quantization is complementary to our method and that the size and complexity of the networks we find can be further decreased using quantization. We believe that our work provides a way to automatically create LLMs which can be used on less expensive and more readily available hardware platforms.

\end{abstract}





\section{Introduction}

Recently, large language models (LLMs) have shown remarkable abilities in solving a variety of natural language and complex reasoning tasks \cite{touvron2023llama}, \cite{touvron2023llama1}, \cite{alpaca}, \cite{achiam2023gpt}, \cite{le2023bloom}. In particular, LLaMA2 \cite{touvron2023llama} is a collection of LLMs ranging in size from 7B to 70B parameters and pre-trained on a large corpora of textual data. While pre-trained models like LLaMA2 can be directly deployed for many downstream tasks, certain applications often require additional supervised fine-tuning to improve performance. For example, fine-tuning datasets like \cite{alpaca} which contain instruction-response pairs help in improving the LLM's ability to follow user defined instructions and generate better responses. 

Deploying LLMs on hardware platforms other than high-end GPUs can be extremely challenging due to extensive computational and memory requirements. Popular model compression approaches such as pruning, sparsification and quantization help with alleviating memory bottlenecks. While quantization approaches (e.g., AWQ \cite{lin2024awq}, GPTQ \cite{frantar2023gptq}) have been extremely effective in reducing model-size, structural pruning and sparsification approaches (e.g, LLM-Pruner \cite{ma2023llmpruner} and SliceGPT \cite{ashkboos2024slicegpt}) often require model re-training to recover lost performance. Structural pruning involves identifying and removing non-critical structures (e.g., groups of related neurons) within the LLM while preserving its functional performance. Sparsification is related to removing individual connections (i.e., neurons) based on certain rules or heuristics. In the case of SliceGPT \cite{ashkboos2024slicegpt}, entire rows in the original weight matrices are removed to be replaced with smaller, dense matrices containing fewer computations. Additionally, most of the pruning and sparsification approaches require specialized kernels to realize the actual compression and speedup on different hardware platforms.

An alternative approach is to decrease model size and complexity via Neural Architecture Search (NAS). One-shot NAS approaches often involve training a super-network once and allow for sampling of sub-networks using the weight-sharing principle. However, training these super-networks from scratch, especially for large models can be time-consuming and compute intensive and hence not well explored for LLMs. Techniques like InstaTune \cite{sridhar2023instatune} leverage off-the-shelf pre-trained weights and proposes creating a super-network during the fine-tuning stage instead.

To mitigate the size and complexity problems associated with LLMs, we have made the following contributions.

\begin{enumerate}
\item To the best of our knowledge, we are the first to apply one-shot NAS to efficiently decrease the size and computational complexity of LLMs. We demonstrate that, for certain standard benchmark tasks, LLaMA2-7B is unnecessarily large and complex.
\item We demonstrate a methodology of LLM compression using one-shot NAS which not only outperforms pruning and sparsification techniques but does so without the additional step of recovery fine-tuning that pruning and sparsification often require.
\item We perform a network parameter analysis to better understand the architectural characteristics of networks found by our method. We demonstrate that no single set of architectural heuristics can be applied to multiple standard benchmark tasks.
\item We demonstrate a framework which produces compressed LLMs that can be used "out-of-the-box" without specialized software kernels or hardware. We also show that these networks can be further compressed using standard quantization techniques without any additional modifications.
\end{enumerate}

\section{Methods}

\subsection{Search Method}
\label{sec:search_method}

Our approach to optimizing large language models for diverse hardware platforms and performance requirements involves adapting the InstaTune \cite{sridhar2023instatune} methodology, a novel paradigm in one-shot Neural Architecture Search (NAS). InstaTune extends the traditional fine-tuning process by making the model's architecture elastic, allowing it to explore a broader design space beyond fixed structures. This flexibility enables adaptation to different tasks, datasets, and computational resources. InstaTune enhances NAS by leveraging pre-existing model weights, avoiding the need to train super-networks from scratch. By embedding the NAS process within the fine-tuning phase, InstaTune conserves computational resources while ensuring that the derived sub-networks are specifically optimized for the target task rather than the pre-training objectives.

In this work, the pre-trained LLaMA2-7B model \cite{touvron2023llama} was fine-tuned using the techniques proposed in InstaTune with the Alpaca \cite{alpaca} dataset, resulting in the creation of a super-network and a search space that then can be used in the search process. However, in contrast to InstaTune, we do not perform any strong teacher or super-network-based knowledge distillation on LLaMA2-7B, primarily due to computational and memory constraints. Additionally, while InstaTune computes losses for both the super-network and a randomly sampled sub-network in the same fine-tuning iteration, we instead alternate between the super-network and the randomly sampled sub-network to alleviate memory usage. The super-network in our case is the base LLaMA2-7B model and does not have any additional layers or expanded intermediate sizes for the MLP modules.

Following InstaTune, an evolutionary search framework was employed to refine the architecture in a multi-objective setting, optimizing for both model size and accuracy on a given downstream task. Specifically, we utilized the Lightweight Iterative Neural Architecture Search (LINAS) algorithm \cite{cummings2022hardwareaware}. LINAS combines NSGA-II search with network performance predictors to efficiently identify Pareto-optimal network configurations. It iteratively evaluates sub-networks on real data to inform performance predictors, which are then used to predict the performance of a large number of sub-networks. The most promising ones are selected for further evaluation in the next round of iterations. This process continues for a specified number of evaluations on real data.

\subsection{Search Space}
\label{sec:search_space}

As mentioned in Section \ref{sec:search_method}, we employ the LINAS algorithm \cite{cummings2022hardwareaware} to perform the optimal sub-network search. After fine-tuning the pre-trained LLaMA2-7B model \cite{touvron2023llama} with the Alpaca \cite{alpaca} dataset using InstaTune \cite{sridhar2023instatune}, we defined a set of allowable parameter values to be used during the search for optimal sub-networks. The search space is created by varying the layer count of the entire network and the intermediate size of each MLP module. These parameter values are shown in Table \ref{tab:search_space} and were used for all tasks described in Section \ref{ssec:tasks}, resulting in a search space size of approximately $1.3\times10^{10}$.

\begin{table}
    \begin{center}
        \begin{tabular}{cc}
            \hline
            \textbf{Parameter} & \textbf{Values} \\
            \hline
            Layer Count & 24, 28, 32 \\
            Intermediate Size & 5504, 11008 \\
            \hline
        \end{tabular}
    \caption{Parameter values used during sub-network search for all tasks with LLaMA2-7B fine-tuned on Alpaca.}
    \label{tab:search_space}
    \end{center}
\end{table}

\section{Evaluation}

\subsection{Hyper-Parameters}

 We first obtained the pre-trained weights from https://huggingface.co/meta-llama/Llama-2-7b and then fine-tuned the pre-trained LLaMA2-7B model \cite{touvron2023llama} on the Alpaca \cite{alpaca} data set for 6 epochs with an initial learning rate of $10^{-5}$ and a global batch size of 128. During fine-tuning, we alternate between the super-network and a randomly sampled sub-network after every iteration and update all the weights.
It is also important to note that we did not use any parameter efficient fine-tuning methods like LoRA \cite{hu2021lora} in our approach.

During the search process with LINAS \cite{cummings2022hardwareaware}, we used a population size of 50, crossover probability of 0.9, crossover eta of 15.0, mutation probability of 0.02 and mutation eta of 20.0 for NSGA-II and performed a total of 250 iterations for each task.

\subsection{Tasks}
\label{ssec:tasks}

We evaluated our methods on a set of standard tasks which are commonly used to quantify the performance of different large language models. Each of these tasks were used to quantify the performance of the pre-trained LLaMA2-7B model \cite{touvron2023llama} and are described below.

\subsubsection{AI2 Reasoning Challenge}

The AI2 Reasoning Challenge (ARC) \cite{clark2018think} is a dataset used for multiple-choice question-answering tasks and tests the effectiveness of sub-networks discovered bu our method in common-sense reasoning. This dataset contains unstructured passages related to science and 7787 associated questions or statements that are designed to asses model's ability to comprehend and reason.

The ARC dataset is commonly sorted into two main splits, namely Easy (5197 questions) and Challenge (2590 questions), where the latter contains questions that require more complex reasoning. Results presented in this work include both ARC Challenge and ARC Easy, denoted as ARC-c and ARC-e, respectively.

\subsubsection{Massive Multitask Language Understanding}

The Massive Multitask Language Understanding (MMLU) \cite{hendrycks2021measuring} dataset is used to measure how well the sub-networks found by our methods have acquired knowledge during training and is similar to how humans are evaluated. It contains 57 subjects across a range of areas including STEM, the humanities, law, ethics and social sciences. It tests not only basic knowledge but also problem solving ability with difficulty range of elementary to advanced.

The MMLU dataset contains 14042 different questions with multiple-choice answers each of which have four different possibilities. A given model is then evaluated on its accuracy of choosing the correct answer for all questions.

\subsubsection{TruthfulQA}

The TruthfulQA \cite{lin2022truthfulqa} dataset is used to measure how well the sub-networks found by our method is truthful in generating answers to questions. It contains 817 questions across 38 diverse categories, such as health, law, finance and politics, and it is designed to identify errors that arise from incorrect beliefs and misconceptions.

The TruthfulQA dataset has two main variants: MC1 (single-true) and MC2 (multi-true). In this study, the MC1 variant is used. For each question there are 4-5 answer choices and the model is evaluated on its ability to choose the one right answer.

\subsubsection{WinoGrande}

The WinoGrande \cite{sakaguchi2019winogrande} dataset tests how well language models handle commonsense reasoning. It includes 44000 problems, specifically designed with challenges like pronoun resolution that are challenging to solve by models that depend on word associations or patterns. It was constructed using a mix of crowd-sourcing techniques and the AfLite \cite{le2020adversarial} algorithm, which effectively reduces systematic biases and enhances the dataset's robustness.

\section{Results}

\subsection{Search Analysis}
\label{ssec:search_analysis}

As described in Section \ref{sec:search_method}, we employ the LINAS algorithm \cite{cummings2022hardwareaware} to search for optimal sub-networks of LLaMA2-7B using the search space shown in Table \ref{tab:search_space}. The search is performed for 250 evaluations on each of the tasks described in Section \ref{ssec:tasks}.

\subsubsection{AI2 Reasoning Challenge}

\begin{figure}
    \centering
    \begin{subfigure}{0.49\columnwidth}
        \centering
        \includegraphics[width=\columnwidth]{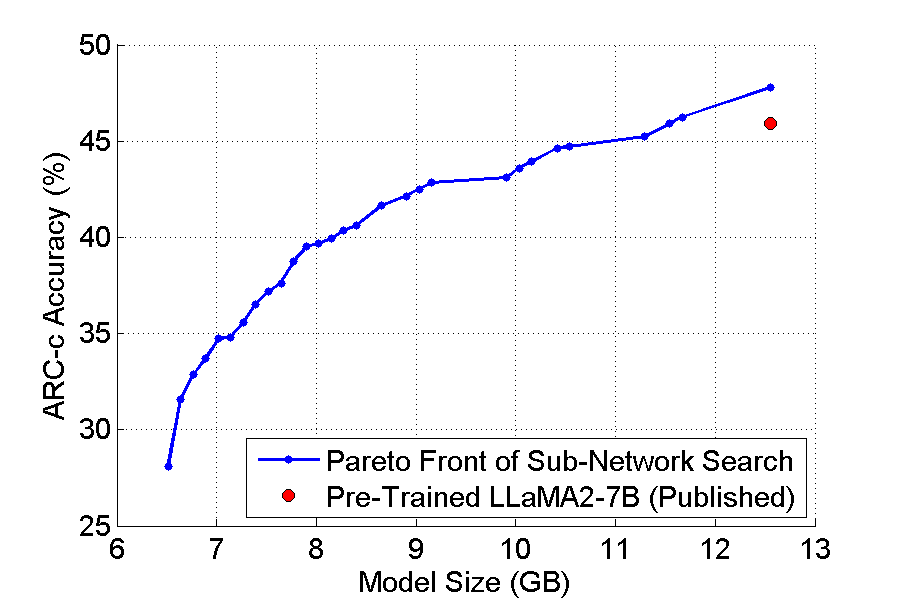}
    \end{subfigure}
    \begin{subfigure}{0.49\columnwidth}
        \centering
        \includegraphics[width=\columnwidth]{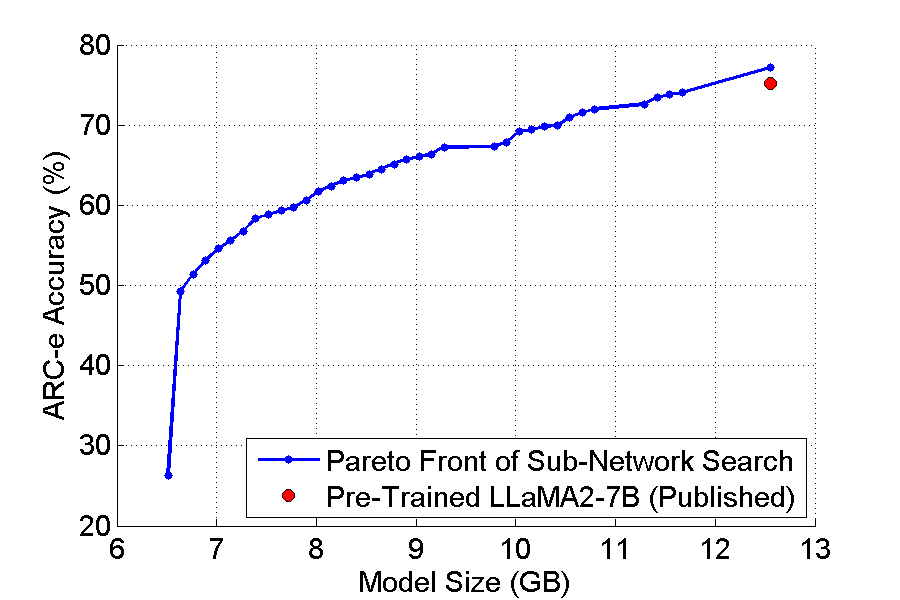}
    \end{subfigure}
    \caption{Pareto fronts after applying our method to search for optimal sub-network architectures in the model size / ARC-c accuracy (left) and model size / ARC-e accuracy (right) objective spaces. The red dot indicates the model size and accuracy of the pre-trained LLaMA2-7B network from \cite{touvron2023llama}.}
    \label{fig:arc_pareto_fronts}
\end{figure}

We first applied our method to the AI2 Reasoning Challenge (ARC) \cite{clark2018think} task and searched in the model size / ARC-c accuracy and model size / ARC-e accuracy objective spaces. Figure \ref{fig:arc_pareto_fronts} shows the Pareto fronts in both of these objective spaces with Alpaca-fine-tuned LLaMA2-7B.

After performing our search, we found several sub-network architectures that provide higher accuracy and / or smaller size. For example, a few sub-networks achieve higher ARC-c accuracy than the pre-trained LLaMA2-7B network shown by the red point. In one case, a sub-network achieves the same accuracy (45.9\%) as the pre-trained LLaMA2-7B network in \cite{touvron2023llama} but is 1.1x smaller while another sub-network with the same size has an accuracy that is 1.9\% higher.

\subsubsection{Massive Multitask Language Understanding}
\label{sec:mmlu_search_analysis}

\begin{figure}
    \centering
    \includegraphics[width=1.0\columnwidth]{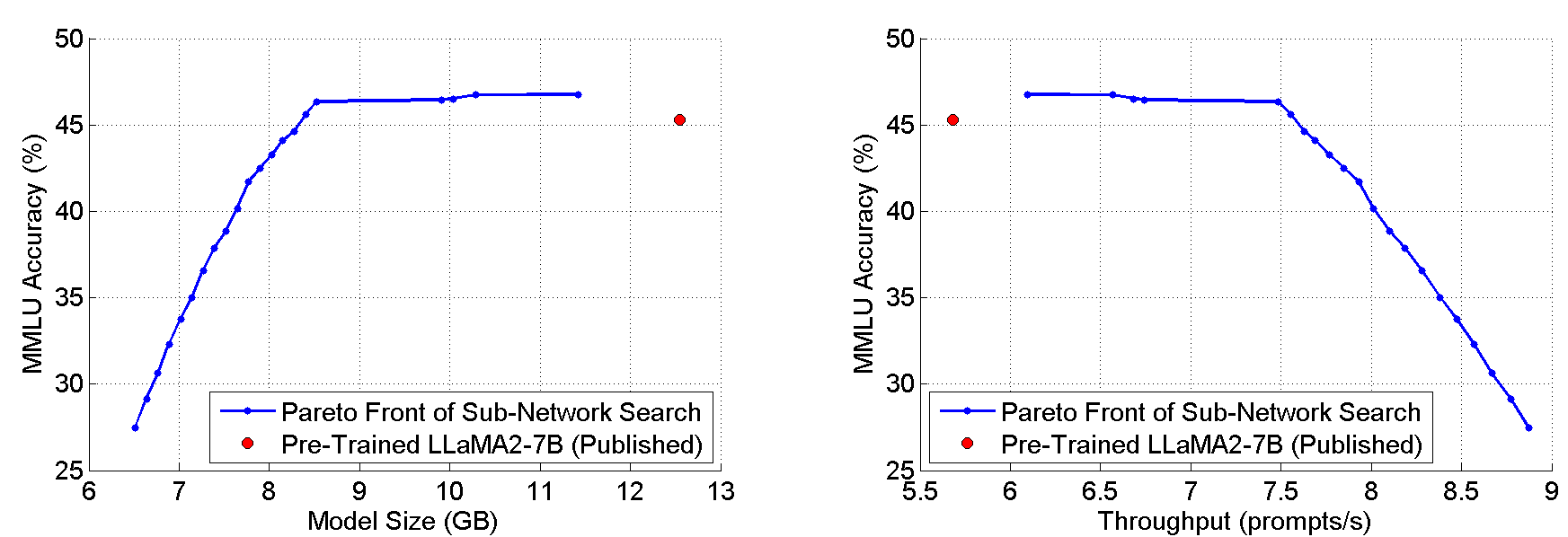}
    \caption{Pareto fronts after applying our method to search for optimal sub-networks with the MMLU task. The left Pareto front is in the model size / MMLU accuracy objective space while the right Pareto front is in the throughput / MMLU accuracy objective space. Throughput is evaluated using a single NVIDIA TitanV GPU with the red dot indicating the model size and accuracy of the pre-trained LLaMA2-7B network from \cite{touvron2023llama}.}
    \label{fig:mmlu_pareto_front}
\end{figure}

We also applied our method to searching for Pareto-optimal sub-networks on the Massive Multitask Language Understanding (MMLU) \cite{hendrycks2021measuring} task using the parameters shown in Table \ref{tab:search_space}. The search was performed in the model size / MMLU accuracy objective space. In addition, we evaluated the throughput of the Pareto-optimal sub-networks found after search to quantify the gain in inference speed. Figure \ref{fig:mmlu_pareto_front} shows the Pareto fronts in both the model size / MMLU accuracy and throughput / MMLU accuracy spaces with Alpaca-fine-tuned LLaMA2-7B.

Within the set of Pareto-optimal sub-networks, several architectures provide clear benefits in both model size and throughput. For example, multiple sub-networks achieve higher MMLU accuracy than the pre-trained LLaMA2-7B network shown by the red point. In one case, a sub-network achieves 1.1\% higher accuracy than the pre-trained LLaMA2-7B network in \cite{touvron2023llama} but is 1.5x smaller and 1.3x faster.

\subsubsection{TruthfulQA}

\begin{figure}[h]
    \centering
    \includegraphics[width=0.5\columnwidth]{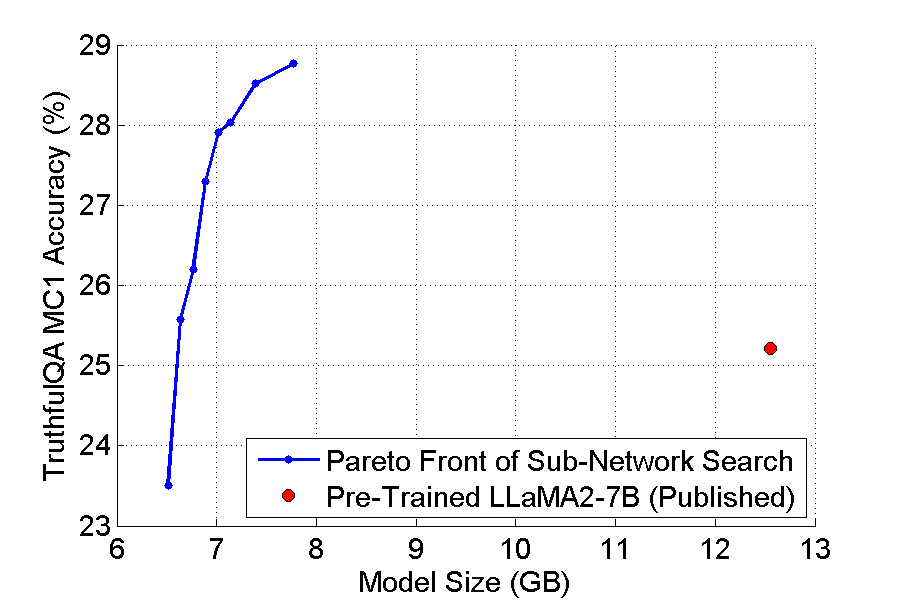}
    \caption{Pareto front after applying our work to Alpaca-fine-tuned LLaMA2-7B in the model size / TruthfulQA MC1 accuracy objective space. The red dot indicates the pre-trained LLaMA2-7B network using the weights from https://huggingface.co/meta-llama/Llama-2-7b.}
    \label{fig:truthfulqa_pareto_front}
\end{figure}

Since the results in \cite{touvron2023llama} did not include the TruthfulQA MC1 \cite{lin2022truthfulqa} task, we first computed the baseline accuracy using the pre-trained weights from https://huggingface.co/meta-llama/Llama-2-7b. We then applied our method to this task and searched in the model size / TruthfulQA accuracy objective space. After our search, we discovered that the pre-trained LLaMA2-7B network is clearly over-parameterized for TruthfulQA MC1. In Figure \ref{fig:truthfulqa_pareto_front}, many network architectures found by our method achieve significantly higher accuracy than the pre-trained LLaMA2-7B network while being significantly smaller. In fact, one network architecture found by our method achieves an accuracy that is 3.6\% higher than the pre-trained LLaMA2-7B network while being 1.6x smaller.

\subsubsection{WinoGrande}

\begin{figure}
    \centering
    \includegraphics[width=0.5\columnwidth]{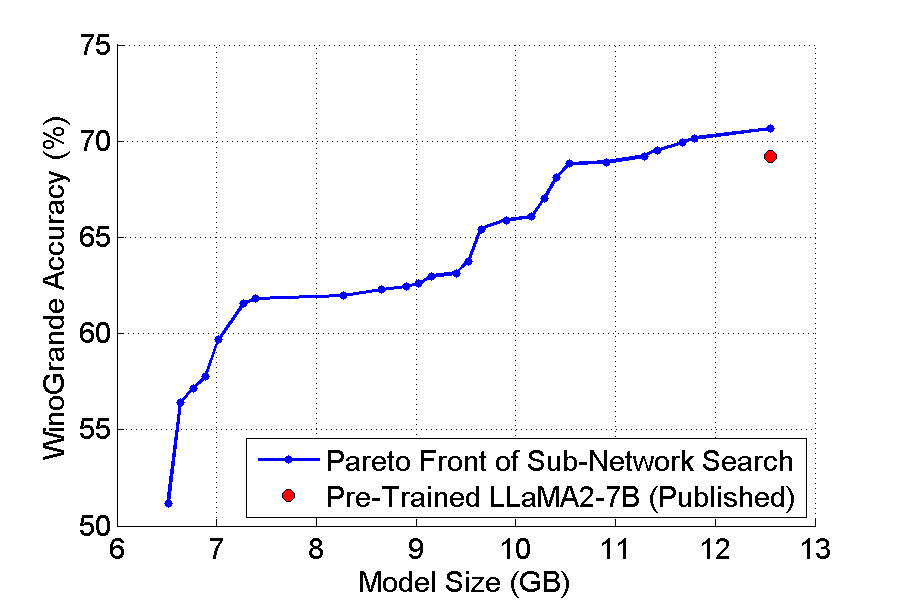}
    \caption{Pareto front after applying our work to Alpaca-fine-tuned LLaMA2-7B in the model size / WinoGrande accuracy objective space. The red dot indicates the model size and accuracy of the pre-trained LLaMA2-7B network from \cite{touvron2023llama}.}
    \label{fig:winogrande_pareto_front}
\end{figure}

When applying our method to the WinoGrande \cite{sakaguchi2019winogrande} task, we found results very similar to those seen for ARC-c. Figure \ref{fig:winogrande_pareto_front} shows the Pareto fronts in the model size / WinoGrande space with Alpaca-fine-tuned LLaMA2-7B.

Our search finds several sub-network architectures which provide higher accuracy and / or smaller size than the pre-trained LLaMA2-7B network in \cite{touvron2023llama}. For example, one sub-network achieves the same accuracy (69.2\%) but is 1.1x smaller while another sub-network with the same size has an accuracy that is 1.4\% higher.

\subsection{Standard Benchmark Performance Summary}

After applying our method to find Pareto-optimal sub-networks for the standard benchmarks, we selected a few notable sub-networks for each task and summarized their standard benchmark performance in Table \ref{tab:benchmark_performance_summary}. Note that the sub-networks highlighted in Table \ref{tab:benchmark_performance_summary} are taken from the Pareto fronts in the Section \ref{ssec:search_analysis} and compared to the standard benchmark performance of LLaMA2-7B.

\begin{table*}[h]
    \begin{center}
        \resizebox{\textwidth}{!} {
            \begin{tabular}{c|cc|cc|cc|cc|cc}
                \hline
                \textbf{Model} & \textbf{Size} & \textbf{ARC-c} & \textbf{Size} & \textbf{ARC-e} & \textbf{Size} & \textbf{MMLU} & \textbf{Size} & \textbf{TruthfulQA} & \textbf{Size} & \textbf{WinoGrande} \\
                \hline
                Pre-Trained & 12.6 GB & 45.9\% & 12.6 GB & 75.2\% & 12.6 GB & 45.3\% & 12.6 GB & 25.2\% & 12.6 GB & 69.2\% \\
                \hline
                \multirow{2}{*}{Ours} & 11.5 GB & 45.9\% & 12.6 GB & 77.2\% & 11.4 GB & 46.8\% & 7.8 GB & 28.8\% & 12.6 GB & 70.6\% \\
                & 10.5 GB & 44.7\% & 11.7 GB & 74.0\% & 8.5 GB & 46.4\% & 6.6 GB & 25.6\% & 11.3 GB & 69.2\% \\
                \hline
            \end{tabular}
        }
    \caption{Selected sub-network performance on standard benchmarks compared to the pre-trained LLaMA2-7B network.}
    \label{tab:benchmark_performance_summary}
    \end{center}
\end{table*}

\subsection{Pruning and Sparsification Performance Comparison}

After applying our method to find Pareto-optimal sub-networks, we selected sub-networks that have the same or similar size as those produced by pruning and sparsification and evaluated them on different tasks. We then compared their standard benchmark performance to LLM-Pruner \cite{ma2023llmpruner} and SliceGPT \cite{ashkboos2024slicegpt} in Tables \ref{tab:pruning_performance_comparison} and \ref{tab:sparsification_performance_comparison}.

\begin{table*}
    \begin{center}
        \resizebox{\textwidth}{!} {
            \begin{tabular}{c|c|c|c|c|c|c|c|c}
                \hline
                \textbf{Model} & \textbf{Pruning Ratio} & \textbf{Parameter Count} & \textbf{Size} & \textbf{ARC-c} & \textbf{ARC-e} & \textbf{MMLU} & \textbf{WinoGrande} & \textbf{Average} \\
                \hline
                \multirow{3}{*}{LLM-Pruner} & 0.25 & 5.4 B & 10.1 GB & 37.5\% & 63.1\% & 25.9\% & 60.2\% & 46.7\% \\
                & 0.32 & 5.0 B & 9.4 GB & 34.0\% & 59.0\% & 23.3\% & 54.4\% & 42.7\% \\
                & 0.38 & 4.7 B & 8.8 GB & 31.7\% & 54.6\% & 25.3\% & 56.2\% & 42.0\% \\
                \hline
                \multirow{3}{*}{Ours} & N/A & 5.4 B & 10.0 GB & 43.6\% & 69.2\% & 46.5\% & 65.6\% & 56.2\% \\
                & N/A & 5.0 B & 9.4 GB & 40.6\% & 67.2\% & 45.5\% & 63.0\% & 54.1\% \\
                & N/A & 4.7 B & 8.8 GB & 41.6\% & 65.1\% & 46.3\% & 61.9\% & 53.7\% \\
                \hline
            \end{tabular}
        }
    \caption{Selected sub-network performance on standard benchmarks compared to LLM-Pruner \cite{ma2023llmpruner} without recovery fine-tuning. All LLM-Pruner accuracies were computed from LLM-Pruner networks generated using https://github.com/horseee/LLM-Pruner since these values were not included in \cite{ma2023llmpruner} for LLaMA2.}
    \label{tab:pruning_performance_comparison}
    \end{center}
\end{table*}

\begin{table*}
    \begin{center}
        \resizebox{\textwidth}{!} {
            \begin{tabular}{c|c|c|c|c|c|c|c|c|c}
                \hline
                \textbf{Model} & \textbf{RFT?} & \textbf{Sparsity} & \textbf{Parameter Count} & \textbf{Size} & \textbf{ARC-c} & \textbf{ARC-e} & \textbf{MMLU} & \textbf{WinoGrande} & \textbf{Average} \\
                \hline
                \multirow{6}{*}{SliceGPT} & No & 20\% & 6.1 B & 11.4 GB & 41.2\% & 71.0\% & 39.8\% & 65.5\% & 54.3\% \\
                & No & 25\% & 5.7 B & 10.6 GB & 38.9\% & 67.9\% & 37.4\% & 64.0\% & 52.0\% \\
                & No & 30\% & 5.3 B & 9.9 GB & 37.8\% & 63.4\% & 32.8\% & 59.8\% & 48.5\% \\
                & Yes & 20\% & 6.1 B & 11.4 GB & 45.1\% & 71.7\% & 39.2\% & 65.6\% & 55.4\% \\
                & Yes & 25\% & 5.7 B & 10.6 GB & 42.8\% & 69.2\% & 36.5\% & 63.2\% & 52.9\% \\
                & Yes & 30\% & 5.3 B & 9.9 GB & 40.9\% & 64.1\% & 34.5\% & 61.6\% & 50.4\% \\
                \hline
                \multirow{3}{*}{Ours} & No & N/A & 6.1 B & 11.4 GB & 45.1\% & 73.4\% & 46.8\% & 69.5\% & 58.7\% \\
                & No & N/A & 5.6 B & 10.4 GB & 44.6\% & 69.9\% & 46.7\% & 68.1\% & 57.3\% \\
                & No & N/A & 5.3 B & 9.9 GB & 43.1\% & 67.9\% & 46.5\% & 65.9\% & 55.9\% \\
                \hline
            \end{tabular}
        }
    \caption{Selected sub-network performance on standard benchmarks compared to SliceGPT \cite{ashkboos2024slicegpt}. The \textbf{RFT?} column denotes whether recovery fine-tuning was performed with Alpaca or not. SliceGPT accuracies for MMLU and ARC-e were computed from SliceGPT networks generated using https://github.com/microsoft/TransformerCompression since these values were not included in \cite{ashkboos2024slicegpt}. All other SliceGPT accuracies were taken from Tables 8 and 10 in \cite{ashkboos2024slicegpt}.}
    \label{tab:sparsification_performance_comparison}
    \end{center}
\end{table*}

In Table \ref{tab:pruning_performance_comparison}, certain sub-networks for ARC-c and MMLU achieve higher accuracy than larger sub-networks found by our solution. The reason for this is that the location of different intermediate sizes within the network architecture may not affect the model size but will affect accuracy. Therefore, a smaller sub-network may achieve higher accuracy because it has a better selection of intermediate sizes. For similarly-sized sub-networks found by our method, the accuracy of each task is higher than both LLM-Pruner and SliceGPT. This also applies to the recovery fine-tuned SliceGPT networks shown in Table \ref{tab:sparsification_performance_comparison} even though our method does not require any post-search fine-tuning.

\subsection{Quantization}

\begin{figure}
    \centering
    \begin{subfigure}{0.49\columnwidth}
        \centering
        \includegraphics[width=\columnwidth]{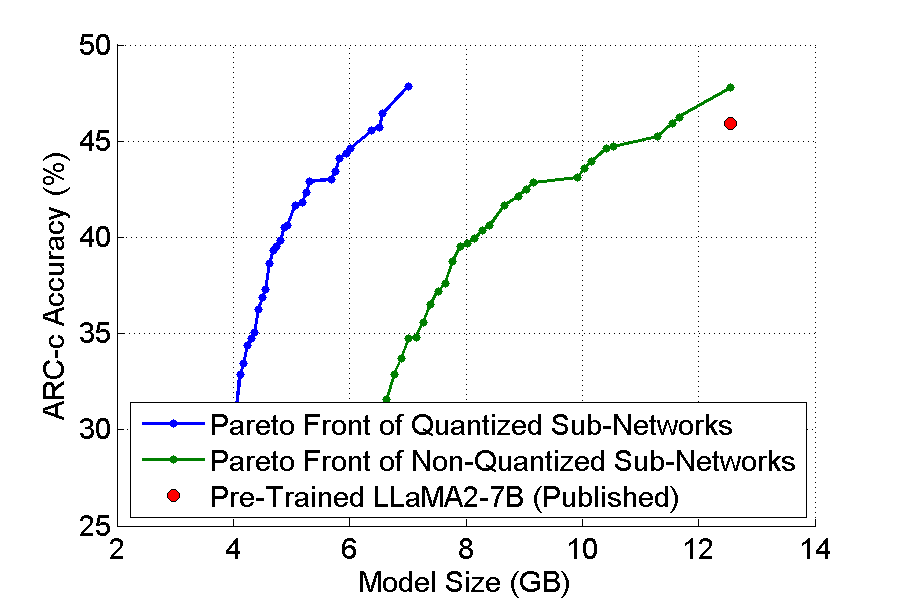}
    \end{subfigure}
    \begin{subfigure}{0.49\columnwidth}
        \centering
        \includegraphics[width=\columnwidth]{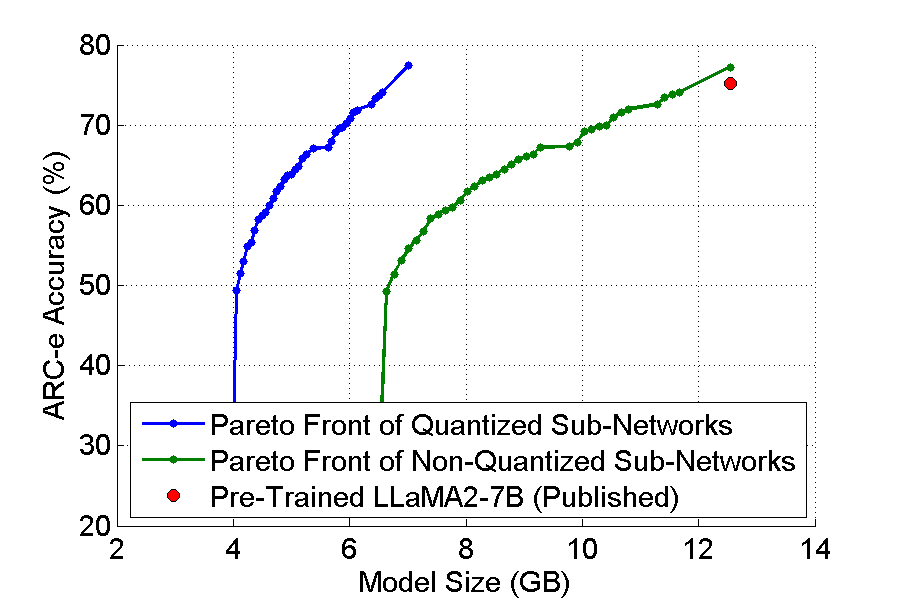}
    \end{subfigure}
    \begin{subfigure}{0.49\columnwidth}
        \centering
        \includegraphics[width=\columnwidth]{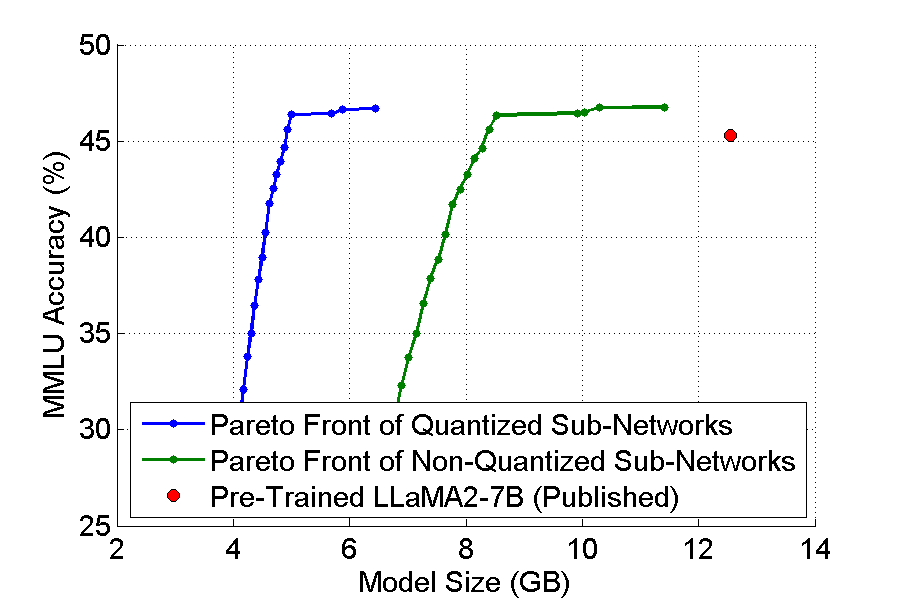}
    \end{subfigure}
    \begin{subfigure}{0.49\columnwidth}
        \centering
        \includegraphics[width=\columnwidth]{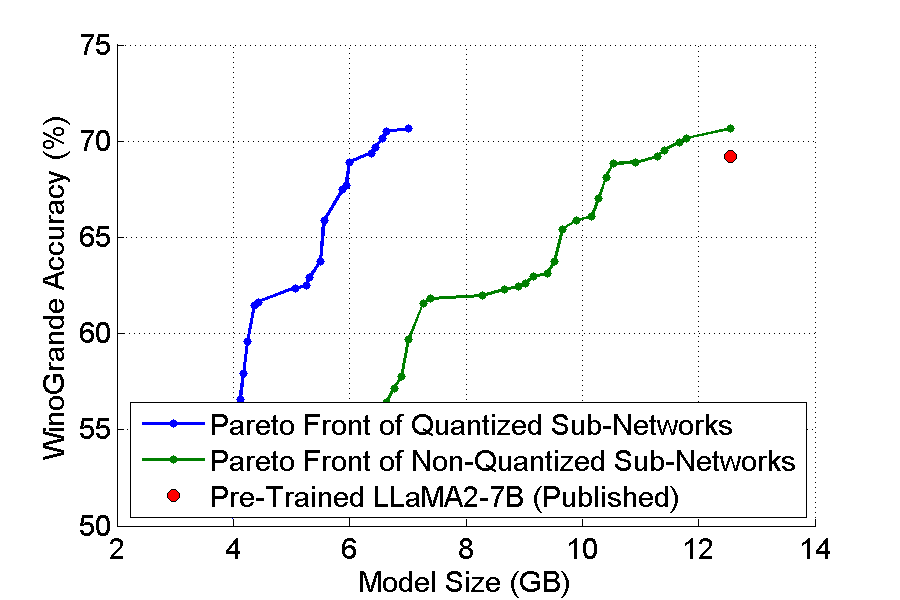}
    \end{subfigure}
    \caption{Pareto fronts before and after applying INT8 quantization to Alpaca-fine-tuned LLaMA2-7B in the model size / accuracy objective spaces. The blue lines are the quantized (INT8) Pareto front while the green lines are original non-quantized (FP16) Pareto front also shown in Figures \ref{fig:arc_pareto_fronts} through \ref{fig:winogrande_pareto_front}. The red dots indicate the model size and accuracy of the pre-trained, non-quantized LLaMA2-7B network from \cite{touvron2023llama}.}
    \label{fig:quantized_pareto_fronts}
\end{figure}

As the sub-networks found by our method do not contain any special operations or other modifications, quantization is simply the straightforward application of quantization using any number of existing techniques. To further improve the performance of the Pareto-optimal sub-networks found using our method, we applied fixed-point (INT8) quantization using \cite{dettmersbnb}. Specifically, we first quantize the Pareto-optimal sub-networks found after search (see Figures \ref{fig:arc_pareto_fronts} through \ref{fig:winogrande_pareto_front}) and then re-evaluate them in the model size / accuracy objective spaces. All linear operations in the decoder layers are quantized to INT8 while the remaining layers (e.g., embedding) are left in FP16. Figure \ref{fig:quantized_pareto_fronts} shows the Pareto fronts both before and after quantizing the sub-networks after search in the model size / accuracy objective spaces.

After applying fixed-point (INT8) quantization to the Pareto-optimal sub-networks, the Pareto fronts have "shifted left" in the objective space. It is clear that quantization has little effect on accuracy but provides a large benefit in model size reduction. In fact, the non-quantized sub-network that achieves 1.1\% higher MMLU accuracy with a 1.5x reduction in model size (see Section \ref{sec:mmlu_search_analysis}) has retained its accuracy after quantization but is now 2.5x smaller.

\begin{table*}
    \begin{center}
        \resizebox{\textwidth}{!} {
            \begin{tabular}{c|cc|cc|cc|cc}
                \hline
                \textbf{Model} & \textbf{Size} & \textbf{ARC-c} & \textbf{Size} & \textbf{ARC-e} & \textbf{Size} & \textbf{MMLU} & \textbf{Size} & \textbf{WinoGrande} \\
                \hline
                Pre-Trained Non-Quantized (FP16) & 12.6 GB & 45.9\% & 12.6 GB & 75.2\% & 12.6 GB & 45.3\% & 12.6 GB & 69.2\% \\
                \hline
                \multirow{2}{*}{Ours Non-Quantized (FP16)} & 11.5 GB & 45.9\% & 12.6 GB & 77.2\% & 11.4 GB & 46.8\% & 12.6 GB & 70.6\% \\
                                                           & 10.5 GB & 44.7\% & 11.7 GB & 74.0\% & 8.5 GB & 46.4\% & 11.3 GB & 69.2\% \\
                \hline
                \multirow{2}{*}{Ours Quantized (INT8)} & 6.5 GB & 45.7\% & 7.0 GB & 77.4\% & 6.4 GB & 46.7\% & 7.0 GB & 70.6\% \\
                                                       & 6.0 GB & 44.6\% & 6.6 GB & 74.1\% & 5.0 GB & 46.4\% & 6.4 GB & 69.4\% \\
                \hline
            \end{tabular}
        }
        \caption{Selected non-quantized and quantized sub-network performance on standard benchmark tasks.}
    \label{tab:quantization_performance_comparison}
    \end{center}
\end{table*}

In Table \ref{tab:quantization_performance_comparison}, we compare the non-quantized sub-networks in Table \ref{tab:benchmark_performance_summary} to the same sub-networks after fixed-point (INT8) quantization for the standard benchmark tasks. With nearly no accuracy loss, the quantized sub-networks for MMLU are 2.0x and 2.5x smaller respectively when compared to the non-quantized (FP16), pre-trained LLaMA2-7B network. The quantized networks for ARC-e have slightly \textit{higher} accuracy than their non-quantized counterparts. For WinoGrande, the quantized sub-networks are 1.8x and 2.0x smaller respectively with a small accuracy \textit{gain} of 0.2\% for the smaller network.

\subsection{Search Space Analysis}

\subsubsection{Layer Count}
\label{ssec:layer_count_analysis}

\begin{figure*}
    \centering
    \includegraphics[width=1.0\textwidth]{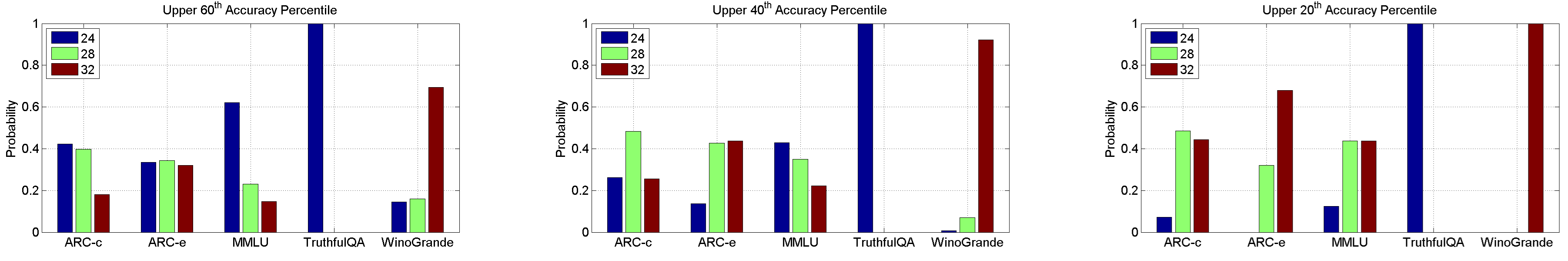}
    \caption{Probabilities that networks selected during the search will have layer count $l\in\{24,28,32\}$ for each task. The probabilities are for sub-networks with accuracies in the upper 60\textsuperscript{th} (left), 40\textsuperscript{th} (center) and 20\textsuperscript{th} (right) percentiles.}
    \label{fig:layer_count_probabilities}
\end{figure*}

To better understand the characteristics of the network architectures found by our method, we performed an analysis of the parameters shown in Table \ref{tab:search_space}. Specifically, we looked at the the probabilities of certain parameter values for all architectures selected during the search. Figure \ref{fig:layer_count_probabilities} shows the probability that a selected network will have layer count $l\in\{24,28,32\}$ for different accuracy percentiles on each of the different tasks.

As expected, a higher layer count tends to result in higher accuracy. However, for certain tasks such as ARC-c, MMLU and TruthfulQA, this is not necessarily the case indicating that LLaMA2-7B is over-parameterized for these tasks. This is particularly evident for TruthfulQA where only 24-layer networks were selected in each of the accuracy percentiles. The over-parameterization for TruthfulQA can also be seen in Figure \ref{fig:truthfulqa_pareto_front} where smaller networks perform significantly better than the much larger LLaMA2-7B network. Only the WinoGrande task chooses 32-layer networks with high probability indicating that this task may be more difficult and require more capacity than the other tasks.

\subsubsection{Intermediate Size}
\label{ssec:intermediate_size_analysis}

\begin{figure*}
    \centering
    \includegraphics[width=1.0\textwidth]{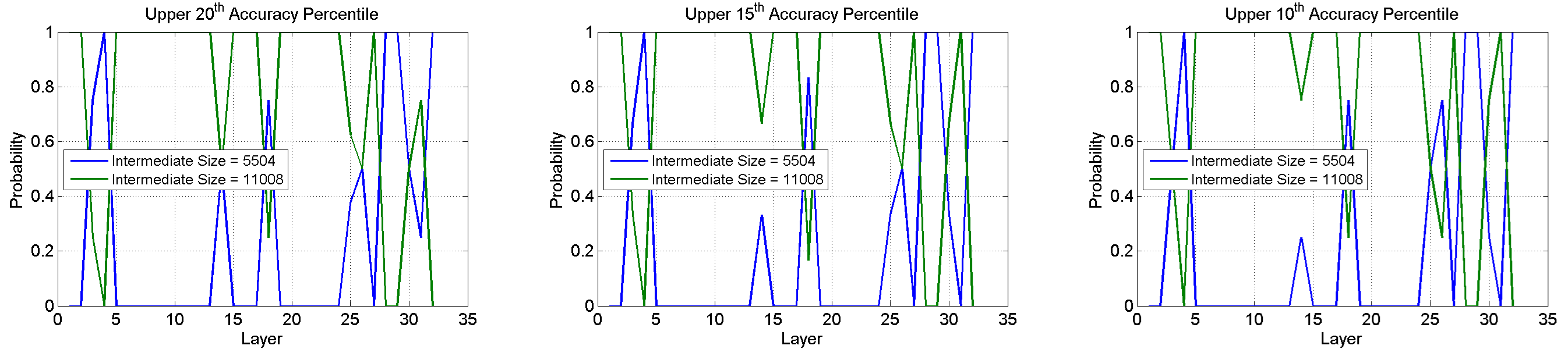}
    \caption{Probabilities of different intermediate sizes after sub-network search in the model size / ARC-c accuracy space with Alpaca-fine-tuned LLaMA2-7B. The probabilities are for 32-layer sub-networks with accuracies in the upper 20\textsuperscript{th} (left), 15\textsuperscript{th} (center) and 10\textsuperscript{th} (right) percentiles.}
    \label{fig:arc_c_intermediate_size_probabilities}
\end{figure*}

After analyzing the layer count probabilities, we looked at which intermediate sizes were chosen during the search. In Figure \ref{fig:arc_c_intermediate_size_probabilities}, we compute the probability that a specific layer will have intermediate size $s\in\{5504,11008\}$ for all 32-layer network architecture selections evaluated on the ARC-c task.

Within these probabilities, the networks which achieve high ARC-c accuracy have a high probability of selecting an intermediate size of 5504 in only certain layers. Specifically, the probability of having intermediate size $s=5504$ in layers 3, 4, 18, 25, 26, 28, 29 and 32 is noticeably higher than $s=11008$. Evaluating the network architecture which has intermediate size $s=5504$ in only layers 3, 4, 18, 25, 26, 28, 29 and 32 and $s=11008$ in all other layers results in an ARC-c accuracy of 45.9\% which matches the accuracy achieved by the pre-trained LLaMA2-7B network in \cite{touvron2023llama} thereby validating what is seen in Figure \ref{fig:arc_c_intermediate_size_probabilities}.

\begin{figure*}
    \centering
    \includegraphics[width=1.0\textwidth]{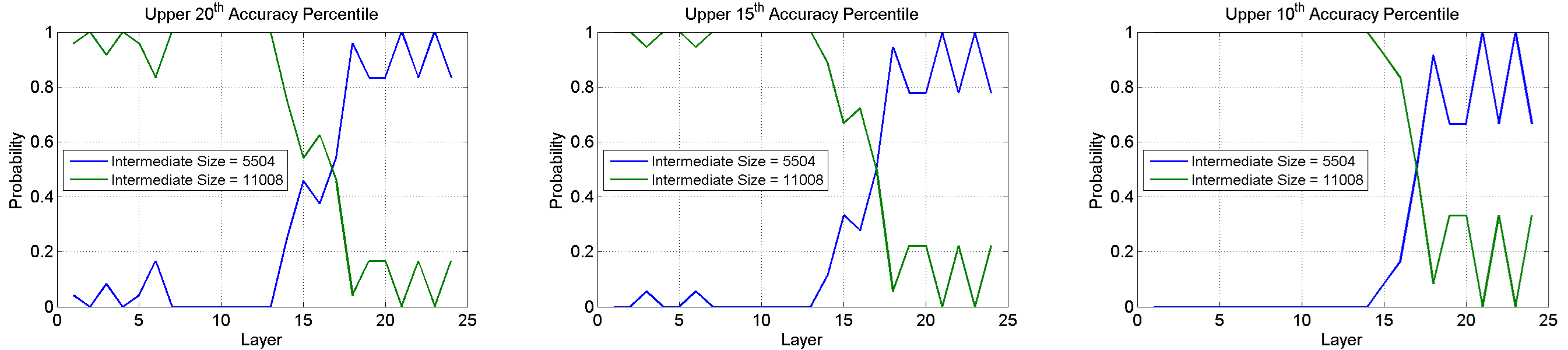}
    \caption{Probabilities of different intermediate sizes after sub-network search in the model size / MMLU accuracy space with Alpaca-fine-tuned LLaMA2-7B. The probabilities are for 24-layer sub-networks with accuracies in the upper 20\textsuperscript{th} (left), 15\textsuperscript{th} (center) and 10\textsuperscript{th} (right) percentiles.}
    \label{fig:mmlu_intermediate_size_probabilities}
\end{figure*}

We perform the same analysis for all 24-layer network architecture selections evaluated on the MMLU task and show the intermediate size probabilities in Figure \ref{fig:mmlu_intermediate_size_probabilities}.

For MMLU, the intermediate size probabilities show a clear pattern where earlier layers (layers 1 through 16) tend to have intermediate size $s=11008$ while layers 18 through 24 tend to have intermediate size $s=5504$. Layer 17 is the "cross-over" point where the probability of the smaller intermediate size is equal to the probability of the larger intermediate size. For additional insight, a 24-layer sub-network with intermediate size $s=11008$ for the first 16 layers and intermediate size $s=5504$ for the last 8 has a size of 8.5 GB and achieves an MMLU accuracy of 46.0\%. This confirms what is shown in Figure \ref{fig:mmlu_intermediate_size_probabilities} since this accuracy is 0.7\% \textit{higher} than the accuracy achieved by the pre-trained LLaMA2-7B network in \cite{touvron2023llama}.

\begin{figure*}
    \centering
    \includegraphics[width=1.0\textwidth]{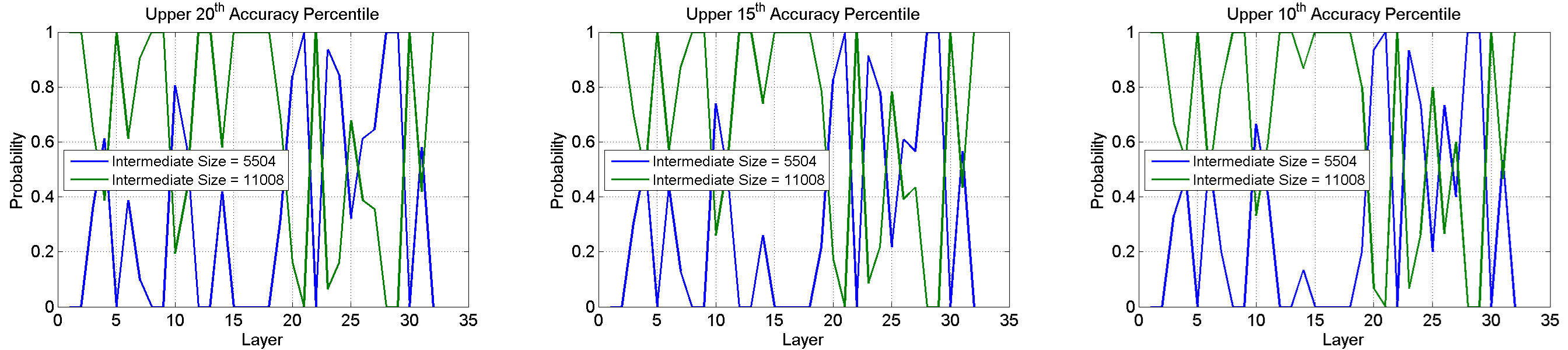}
    \caption{Probabilities of different intermediate sizes after sub-network search in the model size / WinoGrande accuracy space with Alpaca-fine-tuned LLaMA2-7B. The probabilities are for 32-layer sub-networks with accuracies in the upper 20\textsuperscript{th} (left), 15\textsuperscript{th} (center) and 10\textsuperscript{th} (right) percentiles.}
    \label{fig:winogrande_intermediate_size_probabilities}
\end{figure*}

Finally, we show the probabilities for all 32-layer network architecture selections evaluated on the WinoGrande task in Figure \ref{fig:winogrande_intermediate_size_probabilities}.

Although not as clear as the pattern seen with MMLU, the probability pattern for WinoGrande follows ARC-c and shows that only certain, disparate layers have a high probability of selecting an intermediate size of 5504. Specifically, the probability of having intermediate size $s=5504$ in layers 6, 10, 20, 21, 23, 24, 26, 28, 29, and 31 is higher than for $s=11008$. Evaluating the network architecture which has intermediate size $s=5504$ in only those layers and $s=11008$ in all other layers results in a WinoGrande accuracy of 69.2\% which, like ARC-c, matches the accuracy achieved by the pre-trained LLaMA2-7B network in \cite{touvron2023llama}.

Much like the analysis for layer count in Section \ref{ssec:layer_count_analysis}, these results indicate that the full network size is unnecessarily large for certain tasks. For ARC-c, a specific set of 8 layers have a high probability of having intermediate size $s=5504$. This equates to a network with 25\% of all layers having an intermediate size that is half of the original size with \textit{no} accuracy loss compared to the pre-trained LLaMA2-7B network. Similarly, a set of 10 layers have a high probability of having intermediate size $s=5504$ for WinoGrande and a network with that architecture also has no accuracy loss. The results are even more dramatic with MMLU where almost a third of all intermediate sizes in a 24-layer network can be half of the original size while achieving an accuracy \textit{gain}. The combination of fewer layers (24) and smaller intermediate sizes for nearly a third of the remaining layers translates to a 1.5x reduction in model size when compared to the pre-trained LLaMA2-7B network.

Another insight which can be drawn from this analysis is illustrated by the differences in the probability patterns for each task. These differences show that no single set of architectural heuristics can be applied to all tasks. This should be expected since these tasks are chosen as standard benchmarks specifically because they are different from each other and test different capabilities of a given LLM.

\section{Conclusion}

We have proposed an effective method of finding Pareto-optimal network architectures based on LLaMA2-7B. We leverage prior work to generate a super-network fine-tuned on the Alpaca dataset that can be searched using efficient neural architecture search methods. We show that, for certain tasks, the pre-trained LLaMA2-7B network is unnecessarily large and computationally complex. In addition to finding smaller, higher-performing network architectures, our method does so more effectively and efficiently than pruning or sparsification techniques. Our method does not require any specialized software kernels or hardware and works seamlessly with other quantization approaches. As interest in large language models grows, our work provides a way to automatically create networks which can be used on less expensive and more readily available hardware platforms.

\bibliographystyle{unsrt}  
\bibliography{references}

\end{document}